%% file: main.tex
\documentclass{article} 
\usepackage{iclr2021_conference,times}

\input{math_commands.tex}

\usepackage[pdftex]{graphicx}
\usepackage{subcaption}
\usepackage{booktabs} 

\definecolor{mydarkblue}{rgb}{0,0.08,0.45}

\usepackage{hyperref}
\hypersetup{ %
    pdfkeywords={},
    pdfborder=0 0 0,
    pdfpagemode=UseNone,
    colorlinks=true,
    linkcolor=mydarkblue,
    citecolor=mydarkblue,
    filecolor=mydarkblue,
    urlcolor=mydarkblue,
    pdfview=FitH}

\usepackage{url}

\title{Contrastive Embeddings for Neural Architectures}

\iclrfinalcopy

\author{Daniel Hesslow \& Iacopo Poli \\
LightOn\\
\texttt{\{firstname\}@lighton.ai} \\
}

\begin{document}

\maketitle

\begin{abstract}
The performance of algorithms for neural architecture search strongly depends on the parametrization of the search space. We use \textit{contrastive learning} to identify networks across different initializations based on their data Jacobians, and automatically produce the first architecture embeddings independent from the parametrization of the search space. Using our contrastive embeddings, we show that traditional black-box optimization algorithms, without modification, can reach state-of-the-art performance in Neural Architecture Search. As our method provides a unified embedding space, we perform for the first time \textit{transfer learning} between search spaces. Finally, we show the evolution of embeddings during training, motivating future studies into using embeddings at different training stages to gain a deeper understanding of the networks in a search space.
\end{abstract}

\section{Introduction}\label{introduction}
Traditionally, the design of state-of-the-art neural network architectures is informed by domain knowledge and requires a large amount of manual work to find the best hyperparameters. However, automated architecture search methods have recently achieved state-of-the-art results on tasks such as image classification, object detection, semantic segmentation and speech recognition, or even data augmentation and platform-aware optimization~\citep{survey}. 

Neural Architecture Search (NAS) was introduced by \citet{reinforce}, using reinforcement learning. Since then, different search spaces have been developed \citet{nasnet}, and there exists now a set of search spaces that is commonly used to evaluate NAS algorithms \citep{nasbench201, nasbench301, nasbenchnlp}. Different search algorithms have also been conceived: DARTS \citep{darts} relaxes the search space to be continuous to perform architecture optimization by gradient descent, and \textit{aging evolution} \citep{rea} is a genetic algorithm designed for NAS.

Recently, \citet{wo_training} showed that statistics computed on architectures at initialization, before training, can be used to infer which will perform better after training. In particular, they find a heuristic based on samples of the data Jacobians of networks at initialization. Additionally, \citet{wo_training}, is to our knowledge the first instance of a method for NAS that is invariant to the parametrization of the search space, other than random search. 

In parallel, contrastive learning has gathered interest in the computer vision community and produced various state-of-the-art results \citep{moco,simclr,swav,byol}. In contrastive learning, the model learns an informative embedding space through a self-supervised pre-training phase: from the images in a batch, pairs are generated through random transformations, and the model is trained to generate similar (dissimilar) embeddings for similar (dissimilar) images.

In this work, we combine contrastive learning with the idea from \citet{wo_training} of using the Jacobians of networks at initialization, in order to find an embedding space suitable for Neural Architecture Search. We frame our work in the context of the theory presented by \citet{extendedDataJacs}. The embedding space that we generate is invariant to the search space of origin, allowing us to accomplish transfer learning between different search spaces. 

\subsection{Motivations and Contributions}
We design a method to produce architecture embeddings using Contrastive Learning and information available from their initial state. Our technique is capable of generating embeddings independent from the parametrization of the search space, that evolve during training. We leverage these contrastive embeddings in Neural Architecture Search using traditional black-box optimization algorithms. Moreover, since they provide a unified embedding space across search spaces, we exploit them to perform transfer learning between search spaces.
\paragraph{Parametrization-independent embeddings}
NAS methods promise to outperform random search, however the encoding of the architectures must show some structure for the search algorithm to exploit. These embeddings are typically produced by condensing all the parameters used to generate an architecture into a single vector. The method used to generate architectures from a search space thereby implicitly parametrizes it.\\ The parametrization of the search space affects the performance of a NAS algorithm, as noted by \citet{lamts}. However, when performing architecture search, it is not feasible to test multiple different parametrizations of the search space and evaluate which one performs better: once we have started to evaluate networks in a search space, there is no reason to discard previous evaluations. While the current generation of NAS alleviates the need for experts in the design of architectures, now expert knowledge is needed to build and parametrize a search space compatible with the chosen search algorithm, implying that it is exceedingly difficult to outperform a simple random search.\\ We present in Sec. \ref{sec:ourmethod} the first method to create networks embeddings without relying on their parametrization in any search space, through the combination of modern contrastive learning and the theory of data Jacobian matrices for neural architecture search.
\paragraph{Evolution of the embeddings during training} In Section \ref{subsec:embeddingvary}, we show how the embeddings vary during training, noting that the training procedure tends to connect areas with similar final test accuracy together. We hypothesize that this information could enable even more efficient architecture search methods in the future.
\paragraph{Leveraging traditional black-box algorithms} 
Existing methods to generate architecture embeddings rely on metrics from their computational graphs to identify similar architectures, either by explicitly trying to preserve the edit distance in the embedding space, or by leveraging more sophisticated methods from graph representation learning. Our method leverages the information contained in the Data Jacobian Matrix of networks at initialization to train a contrastive model. As such, it can produce embeddings that more meaningfully capture the structure of the search space. As a result, traditional black-box algorithms perform well for architecture search, as shown for NAS-Bench-201 \citep{nasbench201} in Section \ref{subsec:nasbench201}.
\paragraph{Transfer learning between search spaces} Our method provides a unified embedding space, since it does not depend on the parametrization of networks in any search space. We exploit this feature to perform for the first time transfer learning between the two search spaces. In practice, we perform it between the size and the topology spaces in NATS-Bench \citep{natsbench} in Section \ref{subsec:transfer}.

\section{Related Work}

\paragraph{Neural Architecture Search} Previous works have attempted to improve network embeddings: \citet{nasbenchnlp} use \textit{graph2vec} \citep{graph2vec} to find embeddings such that networks with the same computational graph share the same embeddings, and similarly \citet{arch2vec} produce embeddings that are invariant to graph isomorphisms. However, the method differs in that this work trains a variational autoencoder to produce the embeddings. \citet{selfsupNas} uses a contrastive loss to find a low dimensional metric space where the graph edit distances of the original parametrization is approximately preserved.

In the absence of dense sampling, all of these works rely on the prior that the edit distance is a good predictor for relative performance. In contrast, our method, learns to find an embedding space based on intrinsic properties of the architectures. It can therefore discover properties about the architectures which are not present in their graph representation. 

\paragraph{Data Jacobian} Methods based on the Jacobians with respect to the input of trained networks have been shown to provide valuable information for knowledge transfer and distillation \citep{sobolevTraining,knowledgeTransfer}, as well as analysis and regularization of networks \citep{extendedDataJacs}.

\paragraph{Neural Tangent Kernel} The Jacobian of the network with respect to the parameters is computed for inference with \textit{neural tangent kernels} (NTK)~\citep{jacot2018neural}. Using NTK as a proxy for NAS  \citep{park2020towards} underperforms the \textit{neural network Gaussian process} (NNGP) kernel. The NNGP provides an inexpensive signal for predicting if an architecture exceeds median performance, but it is worse than training for a few epochs in predicting the order of the top performing architectures.

\paragraph{Contrastive learning} Different techniques have been developed in contrastive learning. \citet{moco} train a network with a contrastive loss against a memory bank of negative samples produced by a slowly moving average version of itself. \citet{simclr} remove the memory bank and just consider negative samples from within the same minibatch. \citet{byol} remove the negative samples completely but stabilize the training by encoding the positive samples using a momentum encoder. 

\section{Background}

\subsection{Traditional Architecture Embeddings}

A decision tree is created either implicitly or explicitly to sample networks from a search space. To encode an architecture, one records all choices as the decision tree is traversed into a vector, which is then used as the embedding of the architecture. Without any additional knowledge, a NAS algorithm will assume that all choices in the decision tree have an equal importance on the characteristics of an architecture.

\subsection{Data Jacobian}
Extended Data Jacobian matrices are used by \citet{extendedDataJacs} to analyze trained neural networks. We ground our work in their theoretical setting, and introduce the relevant concepts below.

Multi Layer Perceptrons with ReLU activations are locally equivalent to a linear model: the ReLU after a linear layer can be combined into a single linear layer, where each row in the matrix is replaced by zeros if the output pre-activation is negative.

\begin{align*}
    \text{ReLU}(W x) &= \hat{W} x,\quad
    \hat{W}_{ij} = 
    \begin{cases}
    W_{ij} &\text{if } (W x)_j \ge 0\\
    0 &\text{otherwise}
    \end{cases}
\end{align*}

Since matrix multiplication is closed, within a neighborhood where the signs of all neurons pre-activation is constant, the full network can be replaced by a single matrix. This property can be extended to any model whose layers can be rewritten as linear layers, including convolutional layers and average pooling layers. Max pooling layers also retain this property, and can be treated similarly to ReLU. 

Therefore, in a local neighbourhood close to $x$, the full information of a network, $f$, is contained within its \textit{Data Jacobian Matrix} (DJM).
\begin{align*}
 DJM(x) = \frac{\partial f(x)}{ \partial x}
\end{align*}
and within that neighbourhood 
\begin{align*}
    f(x) = DJM(x) x 
\end{align*}

We can evaluate the Data Jacobian Matrix at many different points $x$ to gather information about multiple different neighbourhoods. If we assume the network to have a single output, its DJM is a vector, and we can then stack the DJMs at different points to form the \textit{Extended Data Jacobian Matrix} (EDJM). If a network has multiple outputs we can sum them to get a single output, which we use to calculate the EDJM.

\citet{extendedDataJacs} use the singular values of the EDJM to compute a score, and empirically show that the score is correlated with the depth of the network, and its model capacity. 

\subsection{Contrastive Learning}

\begin{figure}[h]
    \centering
    \includegraphics{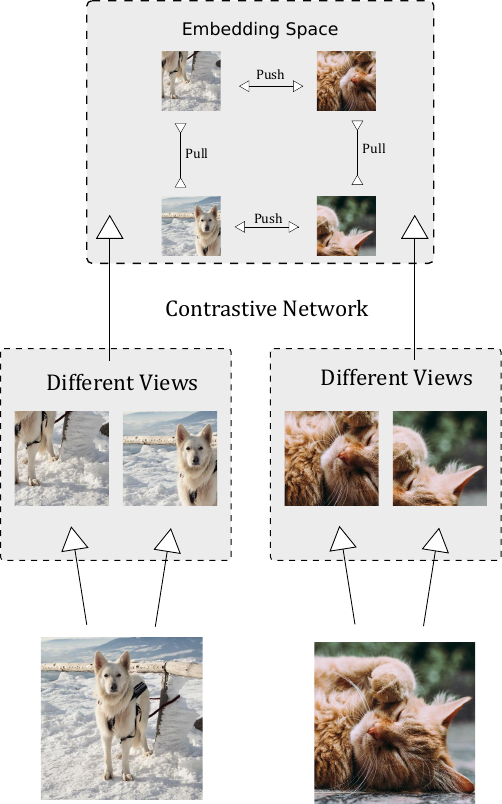}
    \caption{In contrastive learning a network learns to produce similar embeddings for different views of the same picture, while producing dissimilar embeddings for dissimilar pictures.}
    \label{fig:ContrastiveLearning}
\end{figure}

Contrastive learning is a self-supervised method that finds an informative embedding space of the input data useful for downstream tasks. Central to contrastive learning is the concept of a \textit{view} of an object: two different views of the same object are only superficially different, and we should be able to train a network to see past these differences and identify the same underlying object. To this end, a network is trained to map different views of the same object close to each other in the embedding space and, conversely, views of different objects should be far apart from each other, as shown in Figure \ref{fig:ContrastiveLearning}. 

\section{Contrastive Embeddings for Neural Architectures}\label{sec:ourmethod}
We leverage intrinsic properties of the networks to encode them without depending on their parametrization. We must rely on properties of the architectures at initialization, since it is not computationally feasible to train the architectures to obtain their embeddings. At variance with previous work, we develop a method to find such properties automatically, using contrastive learning.

To this effect, we train a network that takes an architecture at initialization as input and produces an embedding at the output. It is desirable that the network has the following two properties:
\newcommand{\norm}[1]{\left\lVert#1\right\rVert}
\begin{itemize}
    \item Different initializations of the same architecture should yield similar embeddings.
    \item Different architectures should yield different embeddings.
\end{itemize}
We can therefore frame our embedding problem as a contrastive learning task: different initializations of the same network will correspond to different views of the same sample in the contrastive learning framework.

\subsection{Our method}
\begin{figure}[h]
    \centering
    \includegraphics{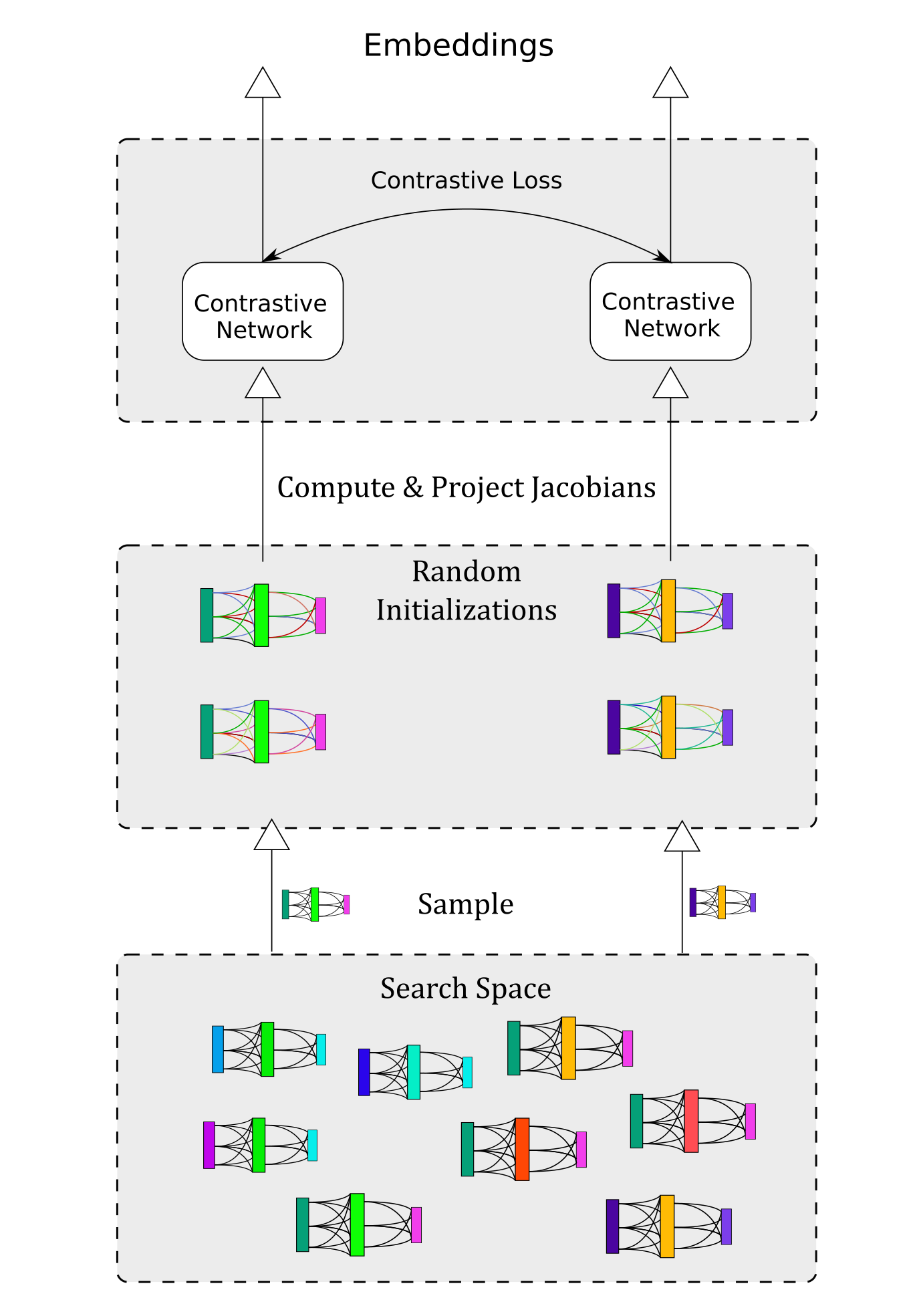} 
    \caption{Illustration of our method for obtaining the network embeddings. We sample architectures from the search space, and form a batch of views with different random initializations. We compute the data Jacobians, project them, and feed them to the contrastive network. The contrastive model learns to generate similar embeddings for networks with similar performance.}
    \label{fig:pipeline}
\end{figure}

Following \citet{wo_training}, we compute the Extended Data Jacobian Matrix (EDJM) of architectures at initialization, and we use a low-rank projection of it as input to our contrastive network, to limit memory requirements. We will refer to the projected version of the EDJM as the \textit{Extended Projected Data Jacobian Matrix} (EPDJM).
\begin{align}
    \text{EPDJM}(X)_i = \phi_X \left ( \frac{\partial \norm{f(X_i)}_1}{ \partial X_i} \right ) 
\end{align}
where $\phi_X$ denotes a projection onto the top-k principal components.
\begin{align}
    X = \begin{bmatrix} U_1 & U_2\end{bmatrix} \begin{bmatrix} \Sigma_1 & 0 \\ 0 & \Sigma_2 \end{bmatrix} V^T, \quad \phi_X(x) = U_1 \Sigma_1 x
\end{align}
The contrastive network is then applied to the EPDJM, and trained using SimCLR \citep{simclr}. Once the Contrastive Network is trained, we can obtain the embeddings of any architecture in the search space, as shown in Figure \ref{fig:pipeline}. The embeddings can then be used with any black box optimization algorithm, we use Sequential Model Based Bayesian Optimization with Gaussian Processes, and use Expected Improvement as the acquisition function. 

\subsection{Implementation}

Since the input of contrastive network is a set of Jacobians at different data points, it is desirable that the architecture of the contrastive network is invariant under permutation of the data points. To this end, we use the simple architecture from \citet{DeepSets}, which encodes each sampled Jacobian matrix with a shared multilayer perceptron (MLP), aggregates them by taking their mean, and finally produces its output with another MLP. 

For the contrastive learning, we use SimCLR with a batch size of 512, and a temperature of 0.1. We project the Jacobians down to a 256-dimensional space. To accelerate the contrastive learning, we precompute the projected Jacobians using four different initializations for each architecture. The computation of the projected Jacobians takes less than 2 hours on a single GPU (NVIDIA RTX 2080Ti). The embedding size is set to 256, and we use a single layer feedforward network for the projection head.

We use the implementation of Gaussian Processes from \citet{gpy2014}, to select new architectures for evaluation by randomly sampling 20 architectures, and choosing the one with the highest expected improvement.
We open source our code, including all hyperparameters\footnote{\scriptsize{\url{https://github.com/lightonai/contrastive-embeddings-for-neural-architectures}}}.

\subsection{Analysis of the Embeddings}

\begin{figure*}[h]
     \centering
     \begin{subfigure}[htbp]{0.32\textwidth}
         \centering
         \includegraphics[width=\textwidth]{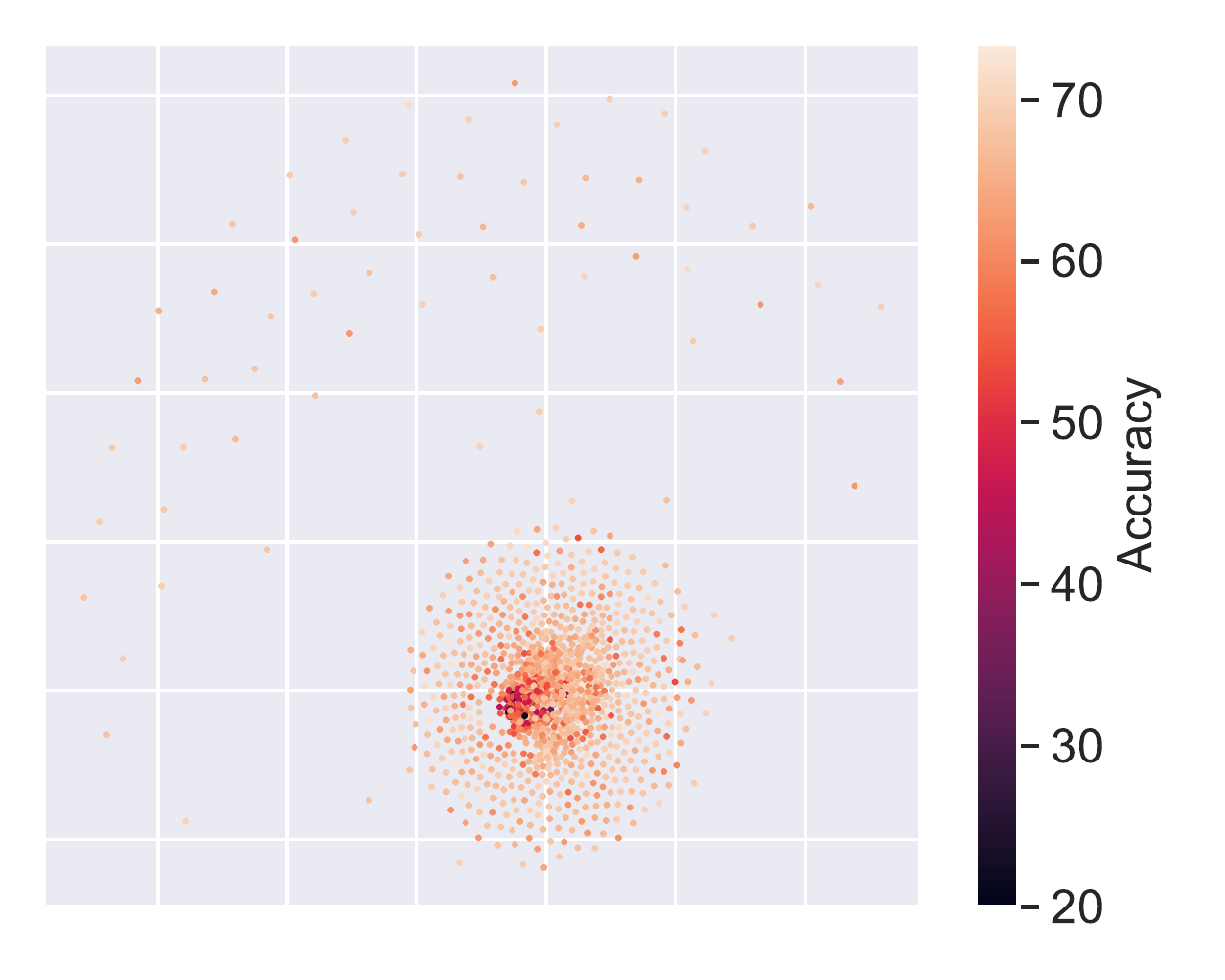}
         \caption{EPDJM}
     \end{subfigure}
     \begin{subfigure}[htbp]{0.32\textwidth}
         \centering
         \includegraphics[width=\textwidth]{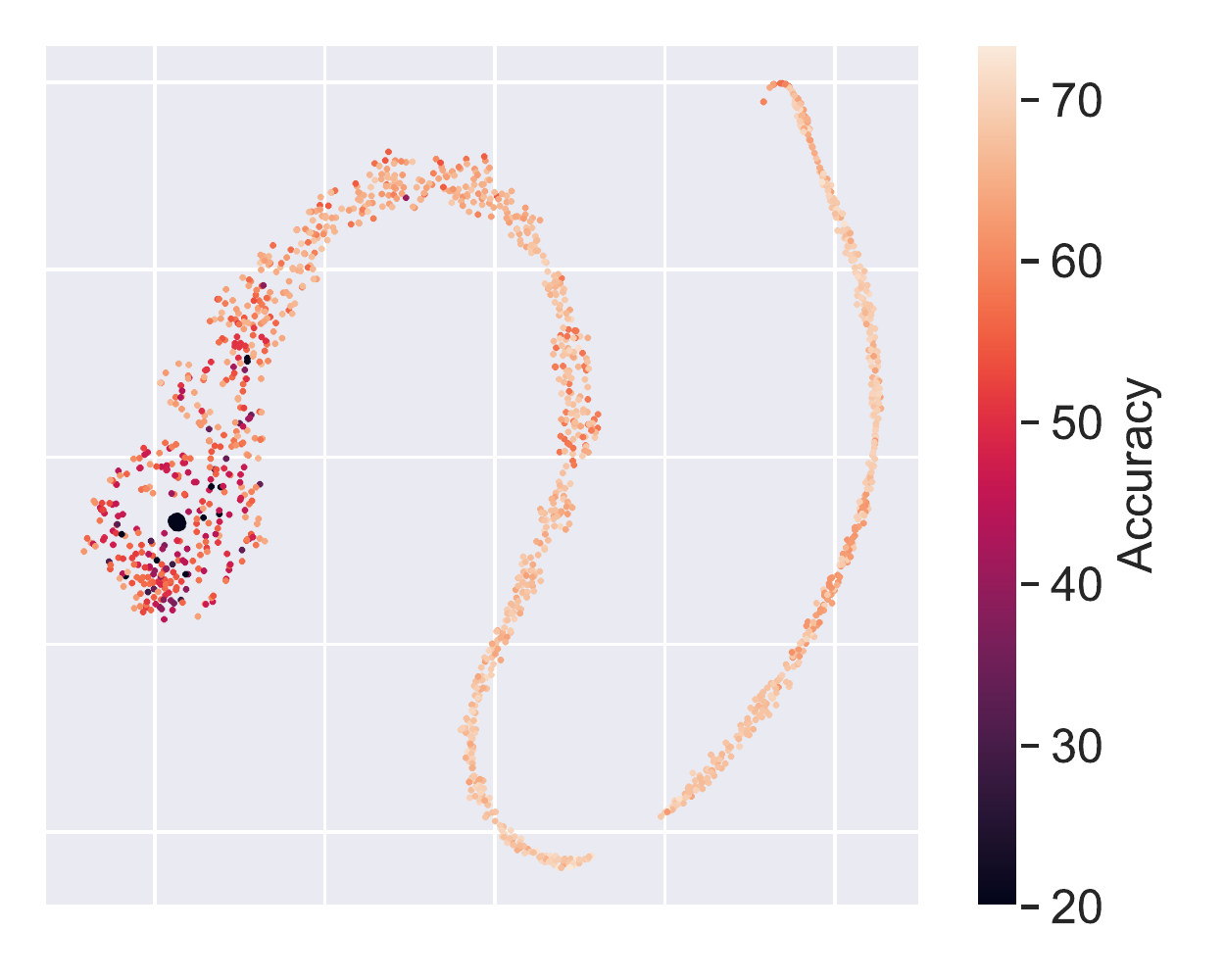}
         \caption{Untrained embedding of the EPDJMs}
     \end{subfigure}
     \begin{subfigure}[htbp]{0.32\textwidth}
         \centering
         \includegraphics[width=\textwidth]{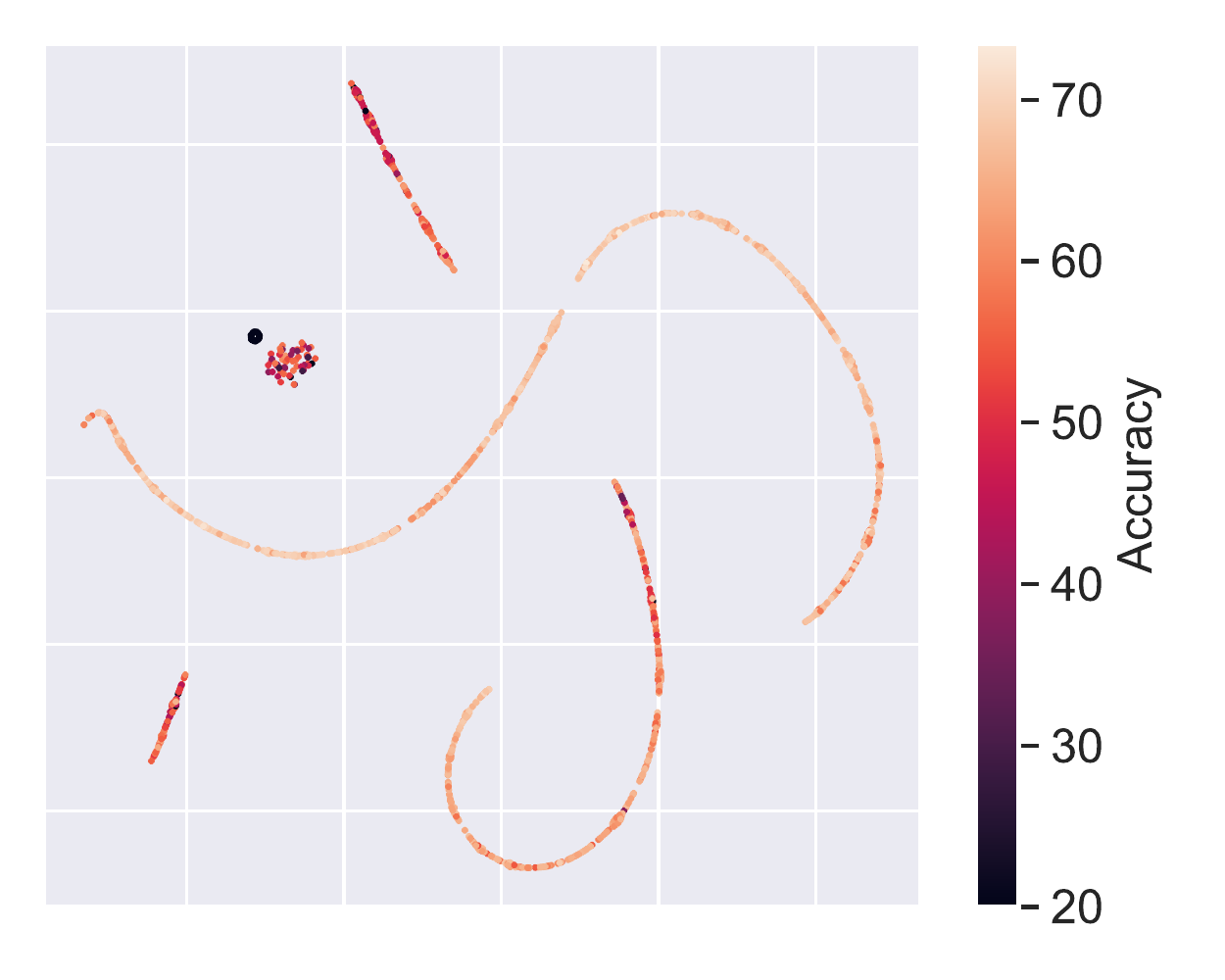}
         \caption{Trained embedding of EPDJMs} \label{fig:five over x}
     \end{subfigure}
        \caption{t-SNE projections of different statistics of 1500 architectures in NAS-Bench-201.}
        \label{fig:tsne}
\end{figure*}

We plot the t-SNE projections \citep{van2008visualizing} at different stages of our method in Figure  \ref{fig:tsne} to analyze the influence of the contrastive learning on the embeddings. We note that the EPDJM alone carries some meaning in the t-SNE space. The contrastive embeddings at initialization of the network already exhibit more evident structure. Finally, the contrastive learning phase produces clean embeddings with little noise, that clearly separate architectures based on performance.

\begin{figure*}[h]
     \centering 
     \begin{subfigure}[htbp]{0.32\textwidth}
         \centering
         \includegraphics[width=\textwidth]{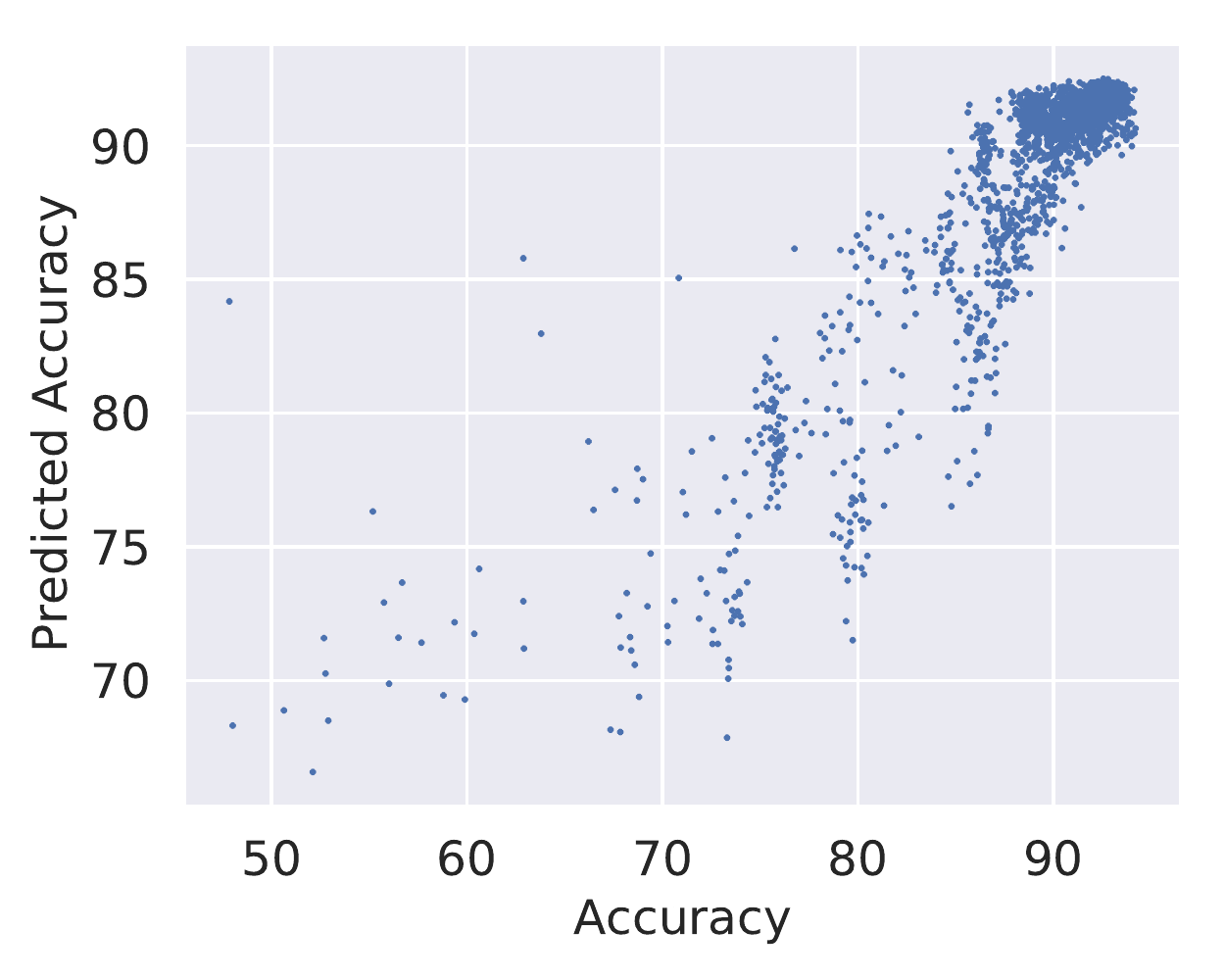}
         \caption{CIFAR-10}
     \end{subfigure}
     \begin{subfigure}[htbp]{0.32\textwidth}
         \centering
         \includegraphics[width=\textwidth]{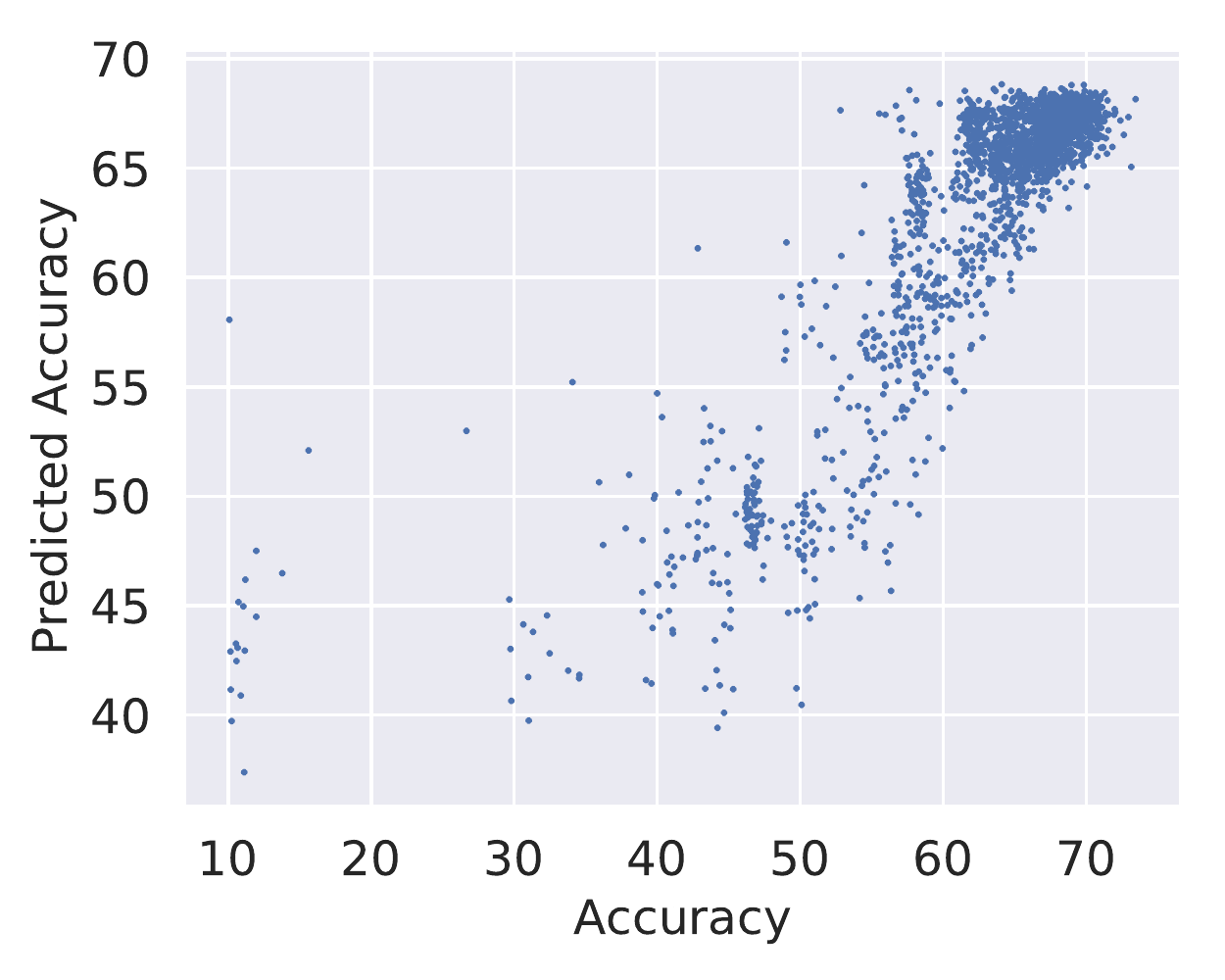}
         \caption{CIFAR-100}
     \end{subfigure}
     \begin{subfigure}[htbp]{0.32\textwidth}
         \centering
         \includegraphics[width=\textwidth]{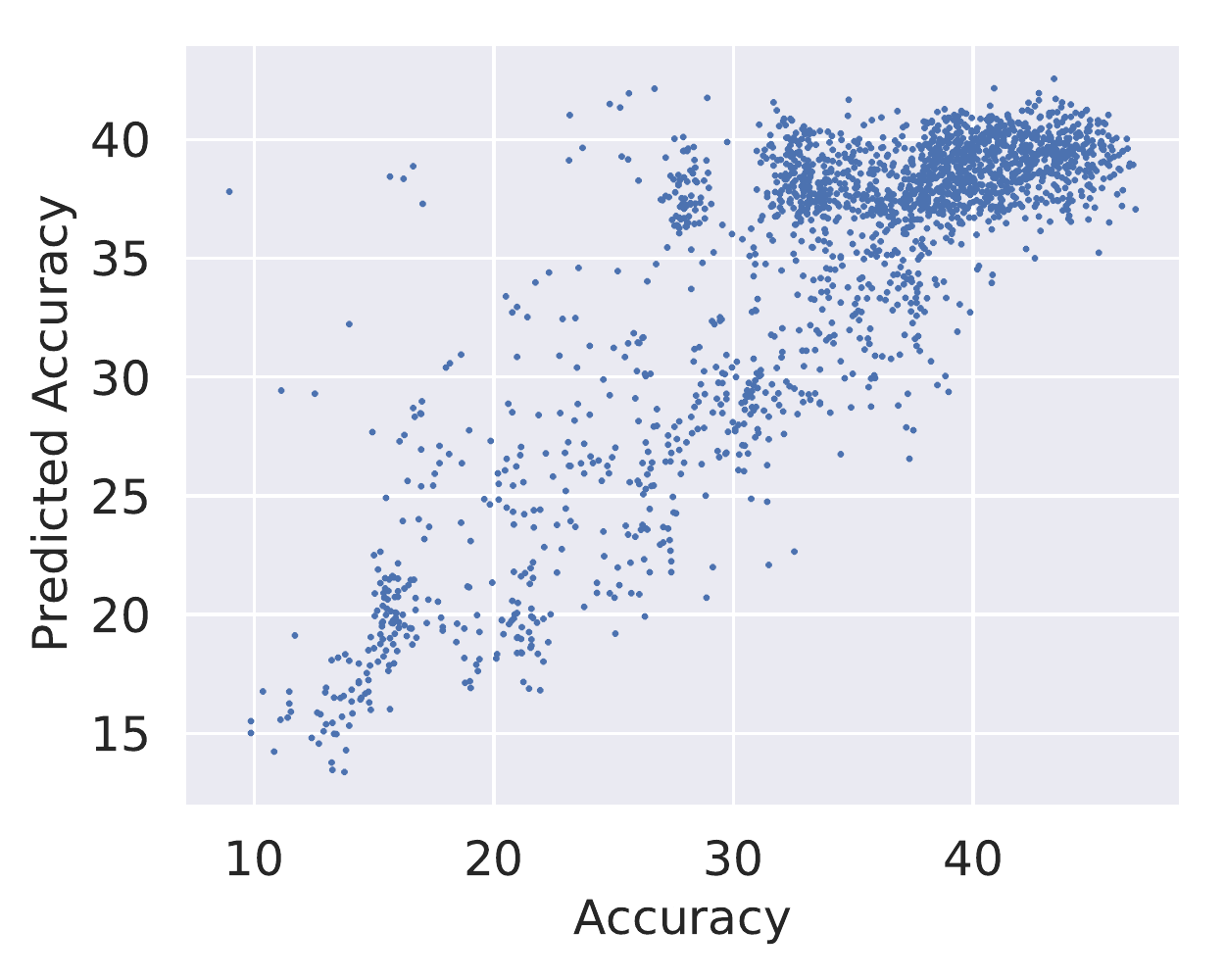}
         \caption{ImageNet16-120}  
     \end{subfigure}
        \caption{Predicted accuracy against actual accuracy. The predictions are produced by LGBM applied on the contrastive embeddings of 500 randomly selected architectures in NAS-Bench-201 \citep{nasbench201}.}
        \label{fig:predictive}
\end{figure*}

Further, we predict the performance of the unseen networks in the search space using LightGBM \citep{lightgbm} with the default hyperparameters, to analyse the predictive power of the embeddings. The results for NAS-Bench-201 \citep{nasbench201} are shown in Figure \ref{fig:predictive} and Table \ref{tab:corr-nasbench}, and indicate that the contrastive embeddings are highly predictive of the performance of the architectures in this search space. 

\begin{table}[h]\caption{Metrics computed on predicted accuracies for the three benchmarks in NAS-Bench-201. This provides a condensed view of Figure \ref{fig:predictive}}
\label{tab:corr-nasbench}
\vskip 0.15in
\begin{center}
\begin{small}
\begin{sc}
\begin{tabular}{lcc}
\toprule
& Correlation & Kendall-$\tau$ \\
\midrule
CIFAR-10 & 0.88 & 0.57 \\  
CIFAR-100 &  0.86 & 0.57 \\
ImageNet16-128 & 0.84 & 0.52 \\
\bottomrule
\end{tabular}
\end{sc}
\end{small}
\end{center}
\vskip -0.1in
\end{table}

\begin{figure}[h]
    \centering
    \includegraphics[width=0.40\textwidth]{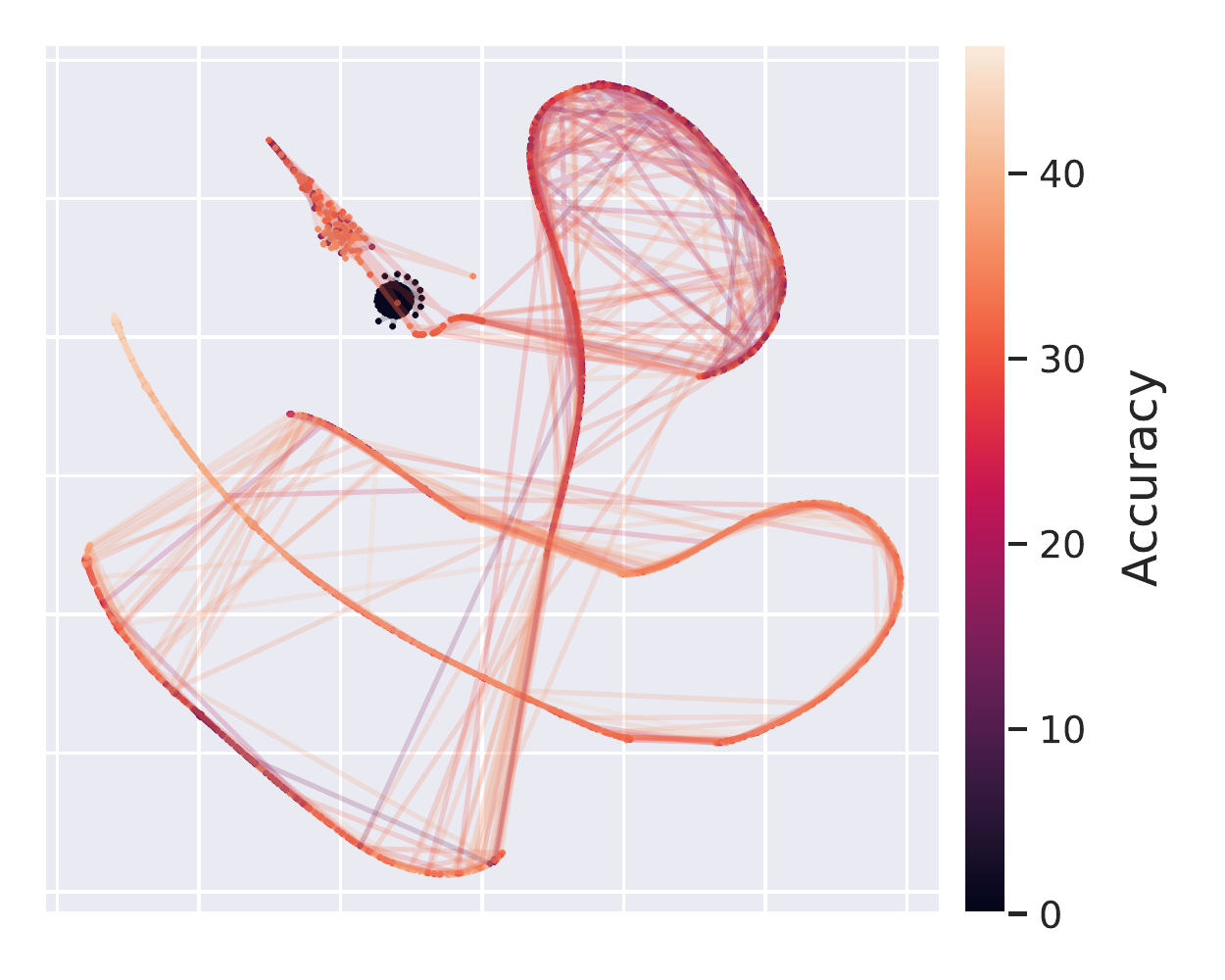}
    \caption{t-SNE projections of movement in embedding space during training of 50 architectures in the NAS-Bench-201 Benchmark. The color of each trajectory represents the final test accuracy of the architecture on the ImageNet16-120 data set, a downsampled version of the traditional ImageNet dataset.}
    \label{fig:optimization_walk}
\end{figure}

\subsection{Embeddings During Optimization}\label{subsec:embeddingvary}
Once the contrastive network is trained, it can produce embedding for architectures at various points during their training. We show the evolution of the embeddings of 100 networks during the 50 first epochs of training in Figure \ref{fig:optimization_walk}: the embeddings vary during the training procedure, potentially enabling future methods to learn more information about the search space for each evaluated network. In particular, the training procedure tends to connect areas with similar final test accuracy.

\section{Architecture Search}
We evaluate our contrastive embeddings on the task of architecture search. We use a Sequential Model Based Optimization (SMBO) with Gaussian Processes \citep{smboBergstra} to guide the search: it is a commonly used method, developed in a different context, and its performance in our setting implies that there is structure in our embeddings that can be easily leveraged.

The covariance function for the Gaussian Process is chosen to be the Matern-52 kernel, which is a stationary kernel that only depends on the Euclidean distance between the architectures in the embedding space. If this can correctly guide our optimization, then the Euclidean distance within the embedding space is a good predictor of relative performance.

Based on the work of \citet{extendedDataJacs}, who construct their score by normalizing the EDJM by the principal singular value, we investigate both the contrastive embeddings produced by the normalized as well as the unnormalized EDJM.

\begin{figure*}[h]
     \centering
     \begin{subfigure}[htbp]{0.3\textwidth}
         \centering
         \includegraphics[width=\textwidth]{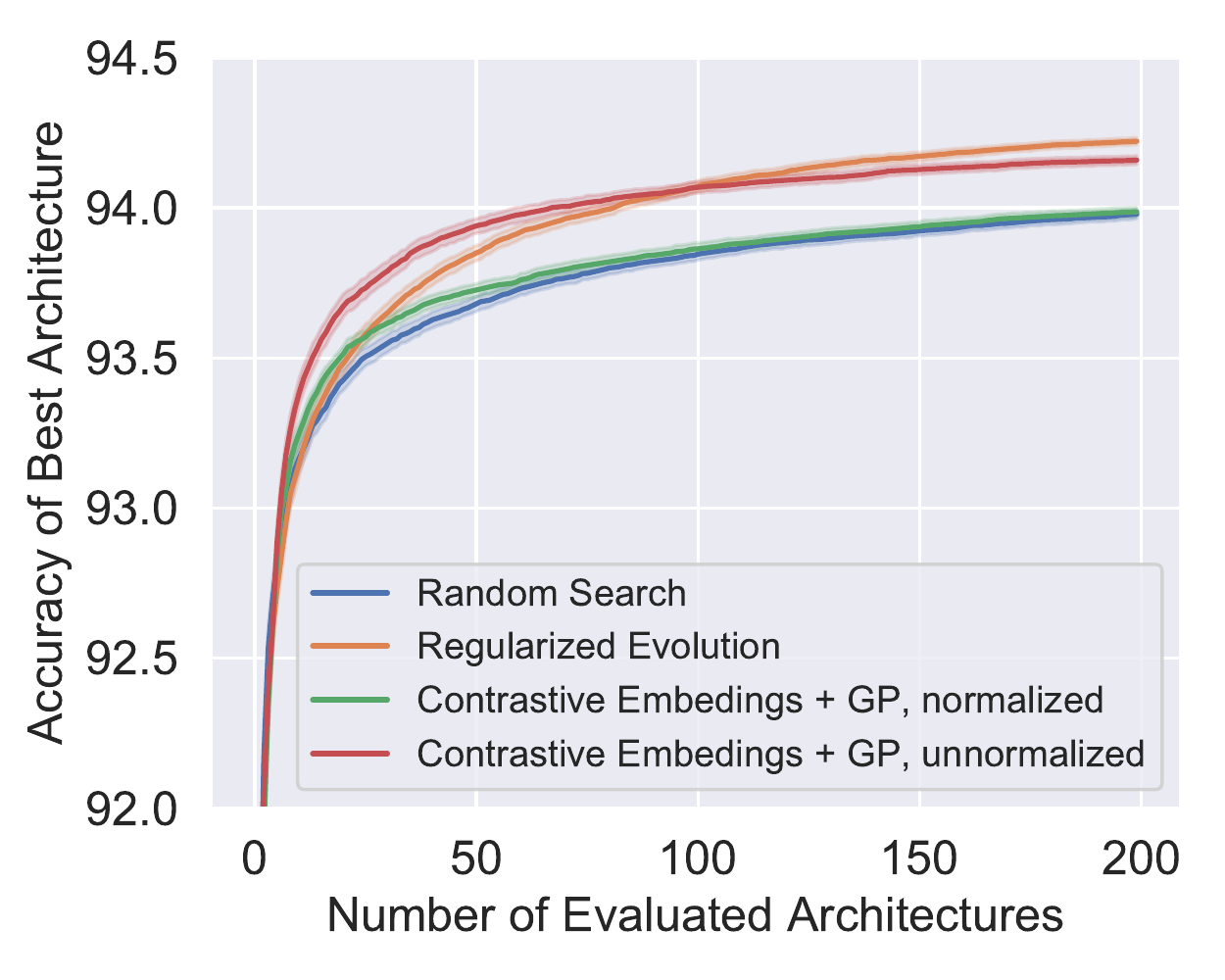}
         \caption{CIFAR-10}
         \label{fig:simulations_a}
     \end{subfigure}
     \begin{subfigure}[htbp]{0.3\textwidth}
         \centering
         \includegraphics[width=\textwidth]{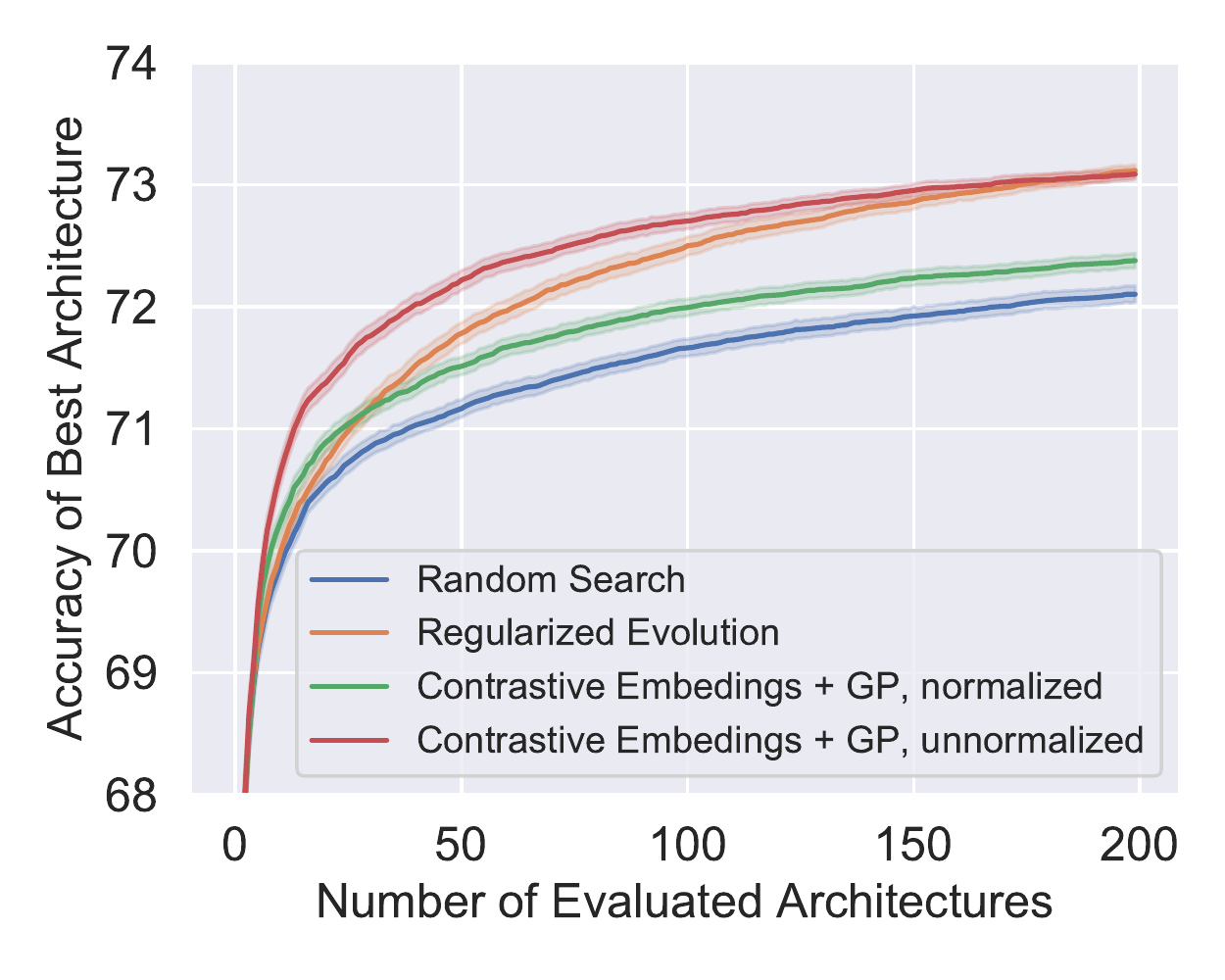}
         \caption{CIFAR-100}
         \label{fig:simulations_b}
     \end{subfigure}
     \begin{subfigure}[htbp]{0.3\textwidth}
         \centering
         \includegraphics[width=\textwidth]{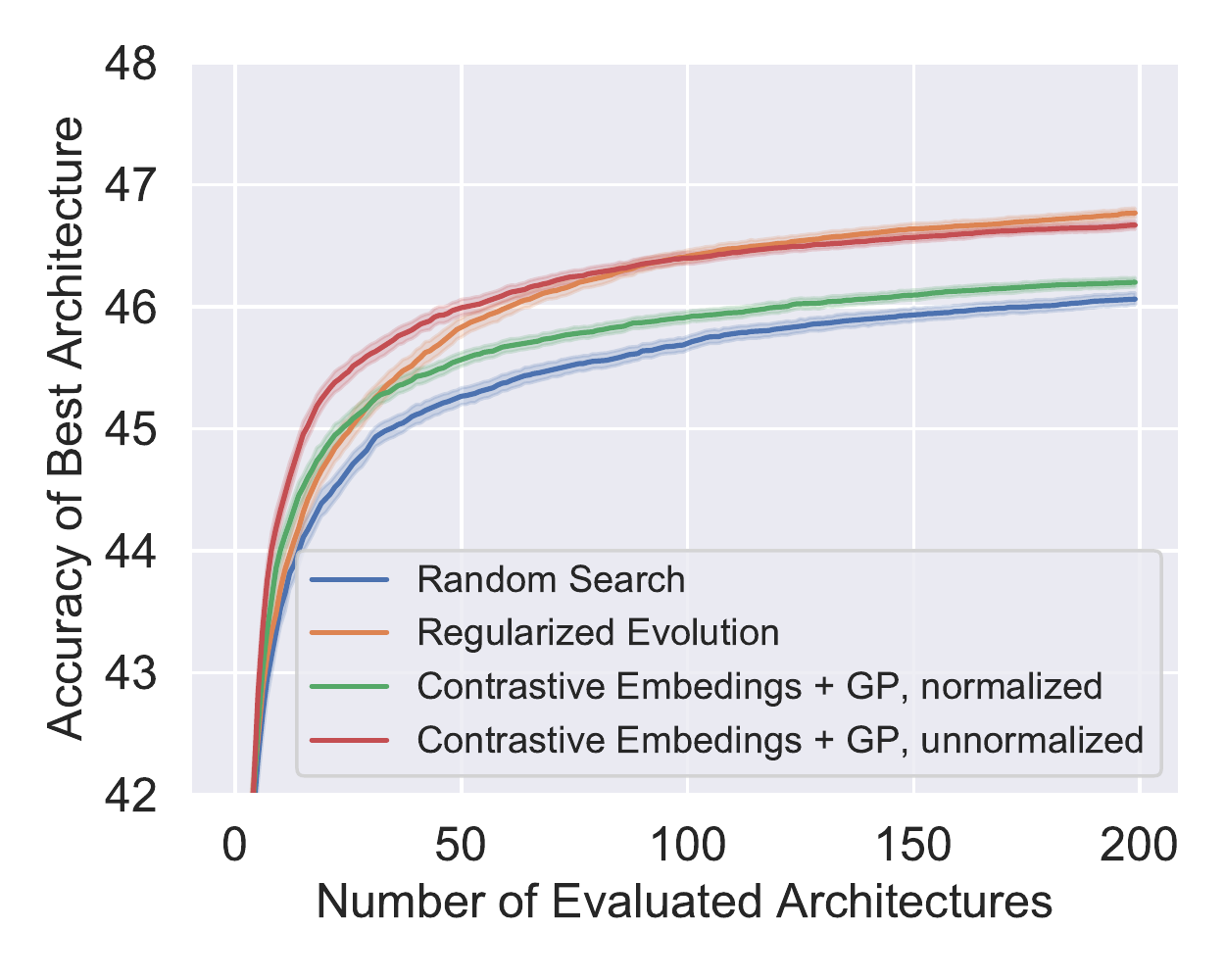}
         \caption{ImageNet16-120}  
         \label{fig:simulations_c}
     \end{subfigure}
        \caption{Search results on NAS-Bench-201
        \citep{nasbench201}
        using the low rank embeddings. We show the results for Contrastive Embeddings produced both by normalizing the principal singular value of the Projected EDJM (normalized) and without (unnormalized).}
        \label{fig:simulations}
\end{figure*}

\subsection{NAS-Bench-201}\label{subsec:nasbench201}

We show the results on the NAS-Bench-201 benchmark \citep{nasbench201} in Figure \ref{fig:simulations}. We notice that the normalization by the principal singular value significantly degrades the performance on this benchmark. However, both versions show a clear improvement over random search, and the unnormalized version also significantly outperforms regularized evolution \citep{rea} when the number of evaluated architectures is small.

We remark that these results require significant manual tuning of the hyperparameters of the optimization algorithm. Nevertheless, they demonstrate that the contrastive embeddings do contain significant structural information. While there are methods to automatically select hyperparameters for Gaussian processes, for example by optimizing the marginal likelihood \citep{gpbook}, we leave this for future work.

\begin{figure*}[h]
     \centering
     \begin{subfigure}[htbp]{0.45\textwidth}
         \centering
         \includegraphics[width=\textwidth]{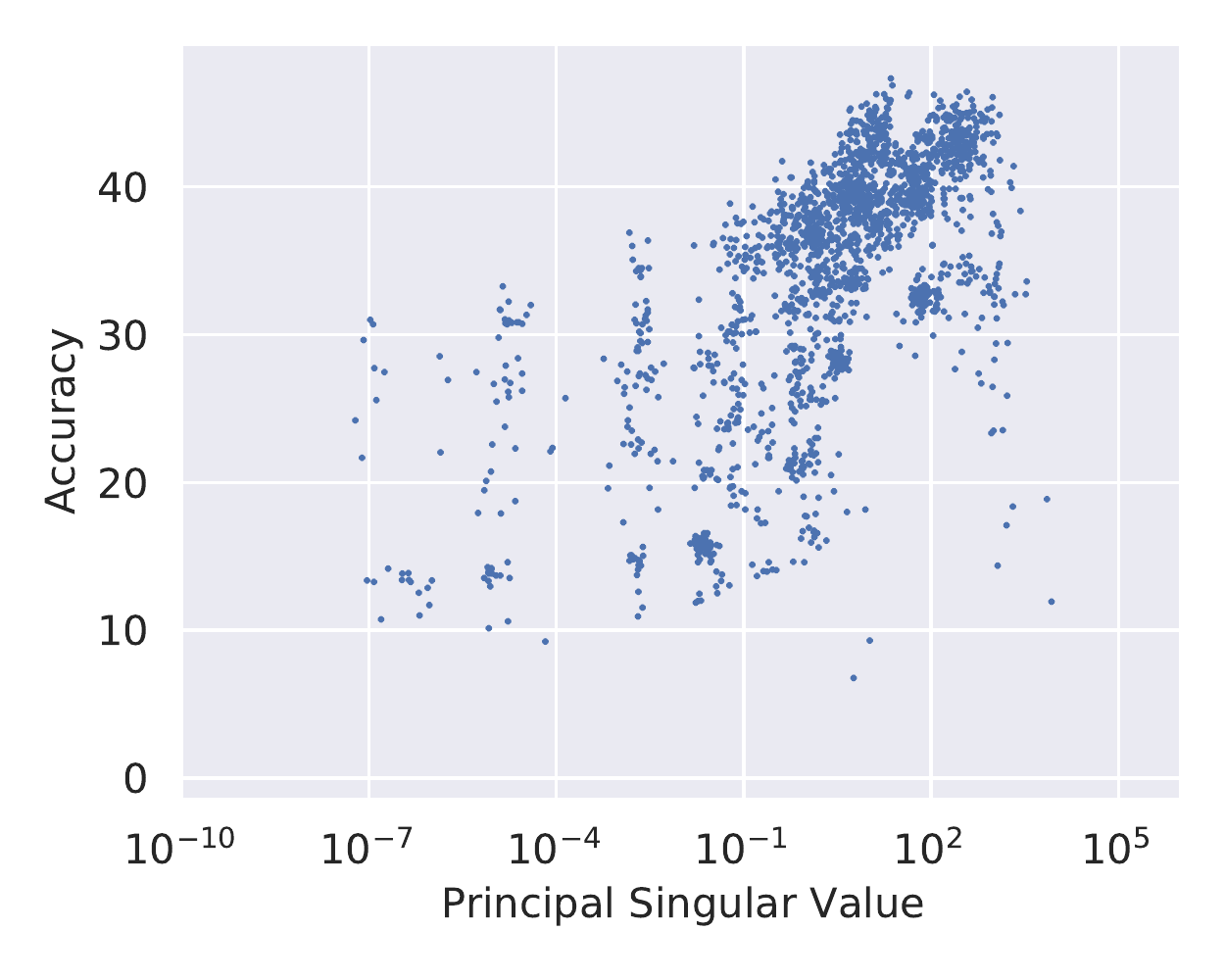}
         \caption{Topology Benchmark}
         \label{fig:psvd_topology}
     \end{subfigure}
     ~
     \begin{subfigure}[htbp]{0.45\textwidth}
         \centering
         \includegraphics[width=\textwidth]{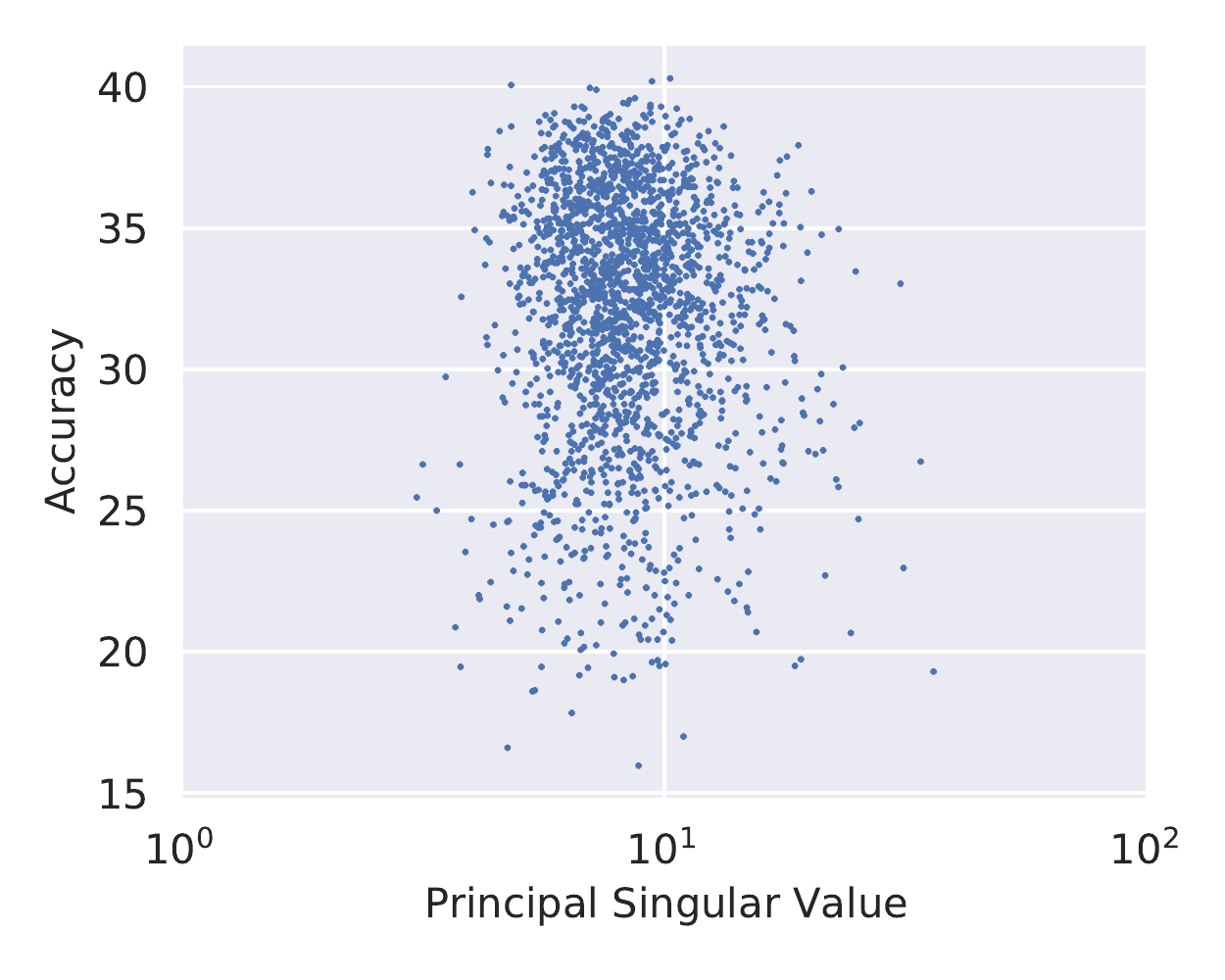}
         \caption{Size Benchmark}
         \label{fig:psvd_size}
     \end{subfigure}
        \caption{The principal singular value of the extended data Jacobian matrix for different architectures in a search space. This quantity is a good predictor of the performance of an architecture on ImageNet16-120 in the topology benchmark, whereas it is not informative for the size benchmark.}
        \label{fig:principal_svd}
\end{figure*}

\begin{figure*}[h]
     \centering
     \begin{subfigure}[htbp]{0.35\textwidth}
         \centering
         \includegraphics[width=\textwidth]{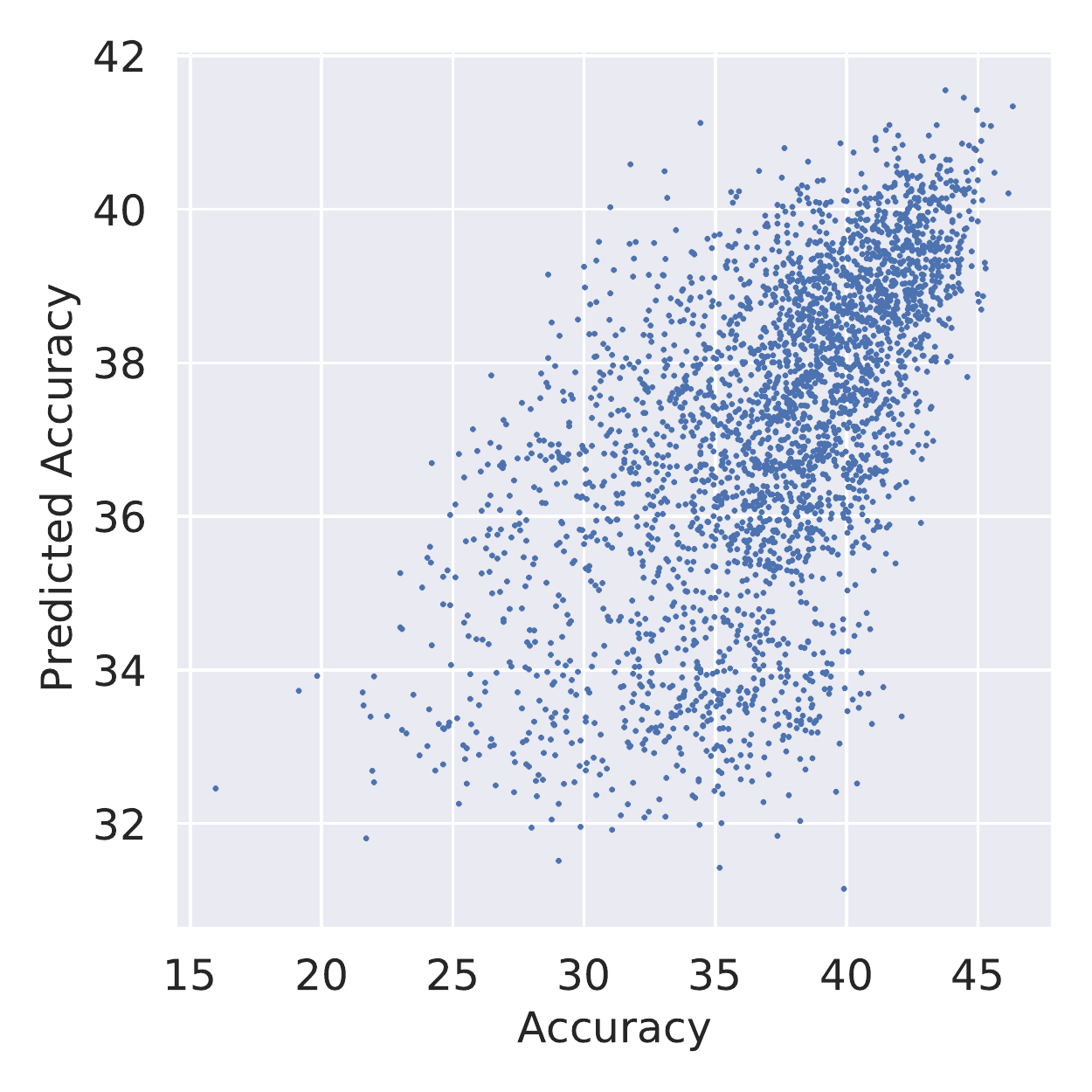}
         
         \caption{size $\rightarrow$ size}
     \end{subfigure}
     ~
     \begin{subfigure}[htbp]{0.35\textwidth}
         \centering
         \includegraphics[width=\textwidth]{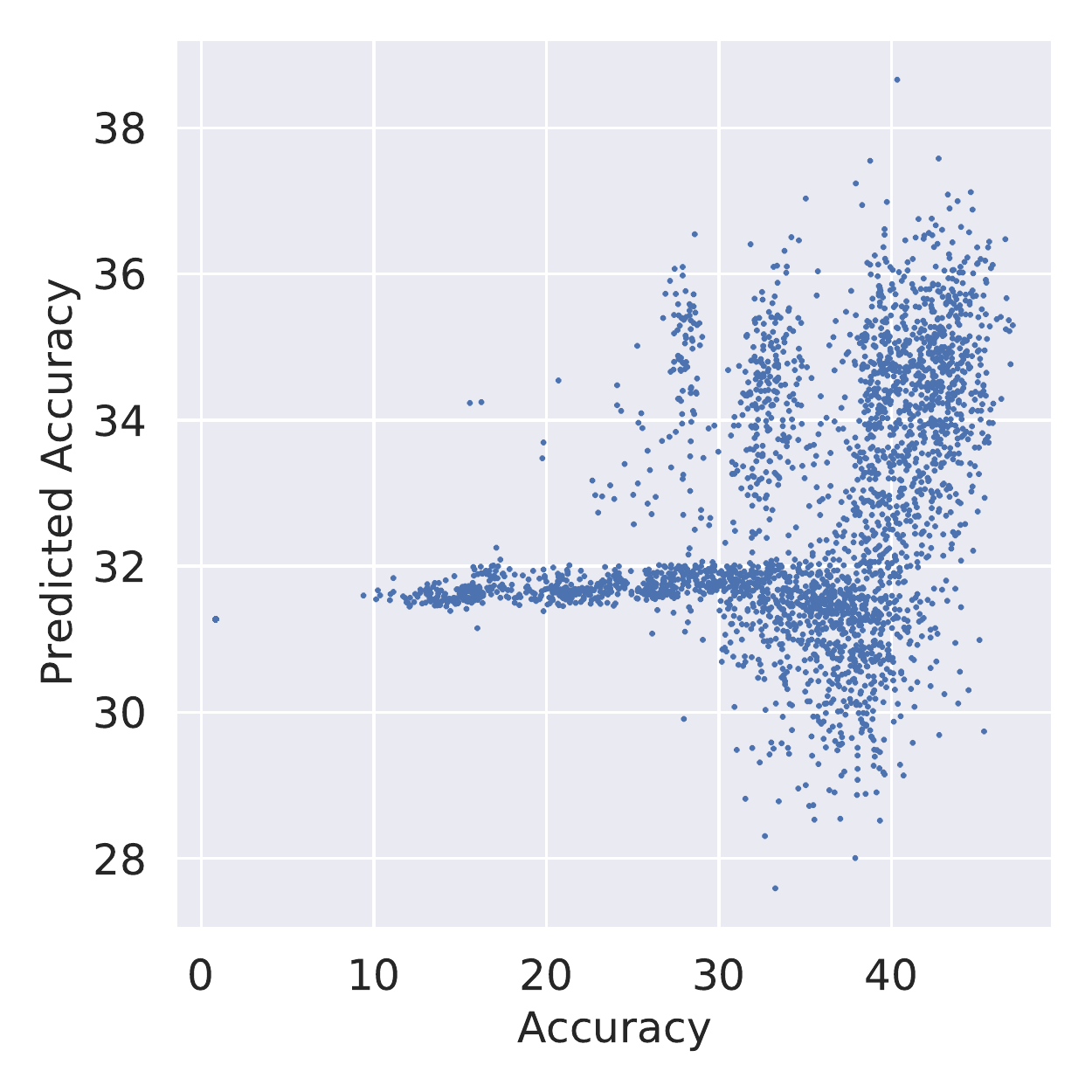}
         \caption{size $\rightarrow$ topology}
     \end{subfigure}
    \\
    \begin{subfigure}[htbp]{0.35\textwidth}
         \centering
         \includegraphics[width=\textwidth]{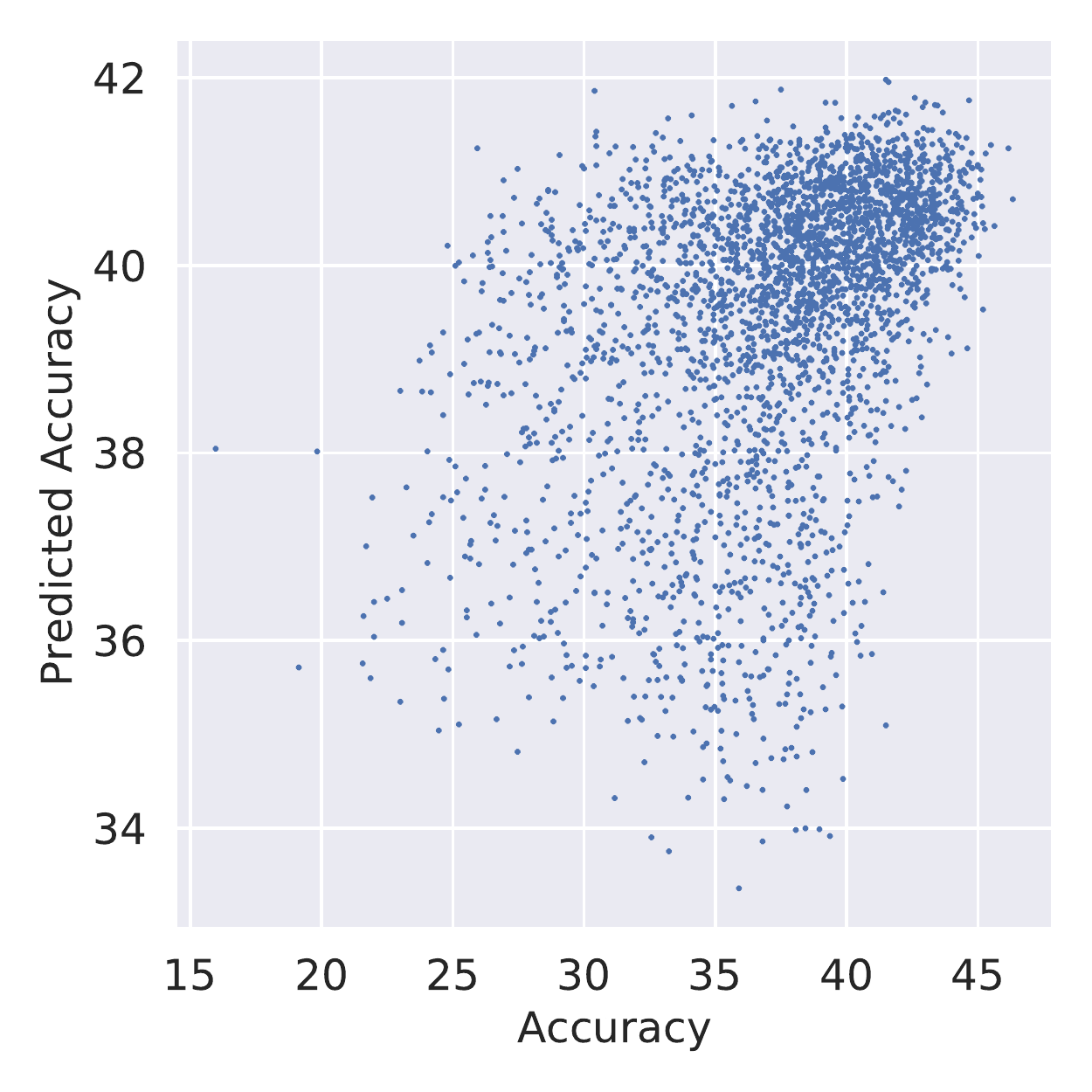}
         \caption{topology $\rightarrow$ size}  
     \end{subfigure}
     ~
     \begin{subfigure}[htbp]{0.35\textwidth}
         \centering
         \includegraphics[width=\textwidth]{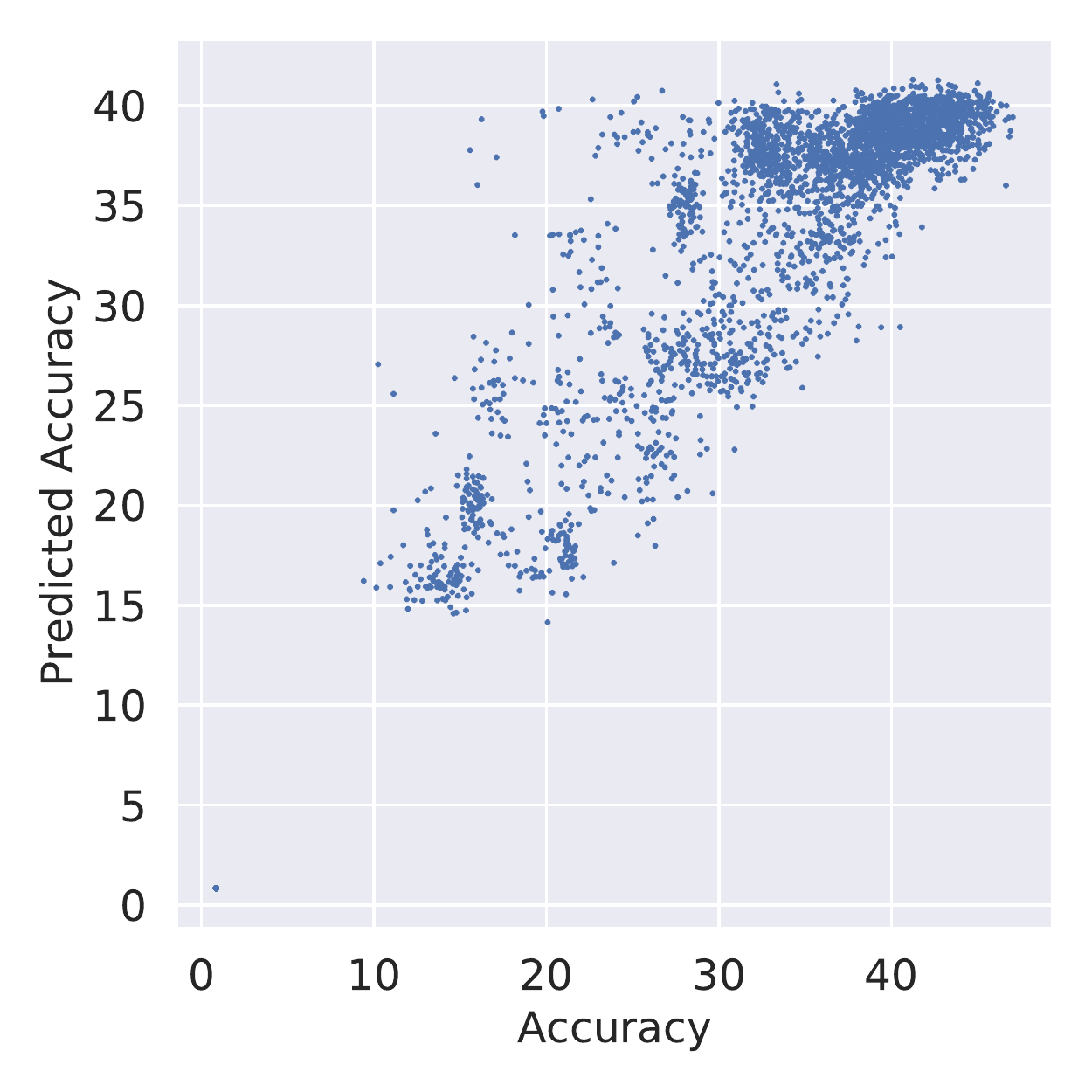}
         \caption{topology $\rightarrow$ topology}  
     \end{subfigure}
    \caption{The transferability of features is evaluated using the two different search spaces that are provided by NATS-Bench: \textit{size} and \textit{topology} \citep{natsbench}. A random forests model is trained on 10000 samples from one benchmark and evaluated on the other. The notation \textit{size}$\rightarrow$\textit{topology} for example means that the model is trained on the size benchmark and evaluated on the topology one. For both \textit{size}$\rightarrow$\textit{topology} and \textit{topology}$\rightarrow$\textit{size} we see a significant correlation between the the predicted accuracies and the actual accuracies without ever having evaluated a single network from the target search space. }
        \label{fig:transfer_learning}
\end{figure*}

\subsection{Transfer Learning}\label{subsec:transfer}
A unique feature of our contrastive embeddings is that they do not depend on any information about the search space used to generated architectures, allowing us to merge the embedding spaces of multiple search spaces into a single one. With such unified embedding space, we perform transfer learning from one search space to another.
 
NATS-Bench \citep{natsbench} consists of two different search spaces: a first one (\textit{topology}) where the topology of architectures is evaluated, and a second (\textit{size}) where the number of filters in different convolutional layers is evaluated. We notice that the two different search spaces have significantly different distributions of the principal singular value. Further, the principal singular value is a good predictor of final performance of the network in the topology benchmark, however it is not in the size benchmark, as shown in Figure \ref{fig:principal_svd}. For this reason, we use the EPDJM normalized by the principal singular value to perform transfer learning. We use random forests \citep{rfs} to predict the accuracy in one search space based on the other, and we display the results in Figure \ref{fig:transfer_learning}.

To evaluate the performance on the transfer learning task we compute two metrics: the Pearson correlation coefficient as well as the Kendall rank correlation the results are shown in \ref{tab:corr-transfer}.  For both \textit{size}$\rightarrow$\textit{topology} and \textit{topology}$\rightarrow$\textit{size}, we see a significant correlation between predicted accuracies and the actual accuracies \textit{without ever having evaluated a single network from the target search space}.

\begin{table}[t]\caption{Metrics computed on predicted accuracies obtained from the transfer learning. A random forest model is trained on a source search space and evaluated on the target search space.  The metrics are reported as \textit{mean} $\pm$ \textit{standard deviation} aggregated over ten runs. The results from the first run can be seen in Figure \ref{fig:transfer_learning}. We remark that \textit{Size}$\rightarrow$\textit{Topology} has a substantially larger standard deviation, which originates in the difficulty of predicting the accuracies of rare architectures with low accuracy in the source search space. In Figure \ref{fig:transfer_learning} these particularly hard to predict architectures form a close to horizontal line. 
}
\label{tab:corr-transfer}
\vskip 0.15in
\begin{center}
\begin{small}
\begin{sc}
\begin{tabular}{r@{\hspace{0em}}lcc}
\toprule
Source&$\rightarrow$Target & Correlation & Kendall-$\tau$ \\
\midrule
Size&$\rightarrow$Size &  0.59 $\pm$ 0.002& 0.39 $\pm$ 0.002 \\
Size&$\rightarrow$Topology & 0.39 $\pm$ 0.051& 0.26 $\pm$ 0.044\\
Topology&$\rightarrow$Size & 0.42 $\pm$ 0.006& 0.31 $\pm$ 0.007\\
Topology&$\rightarrow$Topology & 0.89 $\pm$ 0.000& 0.60 $\pm$ 0.001\\
\bottomrule
\end{tabular}
\end{sc}
\end{small}
\end{center}
\vskip -0.1in
\end{table}

\section{Conclusion and outlooks}
We have developed an end-to-end method to produce embeddings for architectures, using information available from their initial state and contrastive learning, eliminating the need for manual tuning of the search space parametrization. Our analysis of the embeddings clearly shows the advantage introduced by every stage of our pipeline, and our results on Neural Architecture Search using contrastive embeddings are promising. We predict that existing search methods will benefit from further work to improve the encoding stage.

More precisely, by visualizing the t-SNE at different stages of our pipeline, we have shown that the embeddings produced by our technique discover the structure of the search space. We also demonstrated how the embedding space evolves throughout training epochs, connecting regions of the search space with similar final performance, opening the possibility of future work to learn additional information by analyzing these trajectories. 
Moreover, we evaluated how the information content of the embeddings allows us to predict the accuracies of a random subset of architectures from \textit{NAS-Bench-201}. This finding motivated us to employ traditional black-box algorithms to perform architecture search, again on NAS-Bench-201. We remark that even though we used a general-purpose algorithm, not designed specifically for NAS, we reached a clear improvement over random search, and we outperformed Regularized Evolution under a small evaluation budget. These results demonstrate unambiguously that the contrastive encoding learns properties of the networks with predictive power for the performance after training. 

Finally, since the embeddings are independent of the search space, our technique provides a unified embedding space and enables to learn universal properties of the networks. We verified this by performing for the first time \textit{transfer learning across search spaces} in neural architecture search.

Our work highlights a novel direction of work where the focus is not on algorithms for neural architecture search, and instead on improving the embeddings of the neural networks, so that existing methods work better. We wish to inspire further cross-pollination between contrastive learning, black-box optimization and neural architecture search.

\bibliography{main}
\bibliographystyle{iclr2021_conference}

\thispagestyle{empty}
\vfill
\begin{center}
\includegraphics[width = 0.12\textwidth]{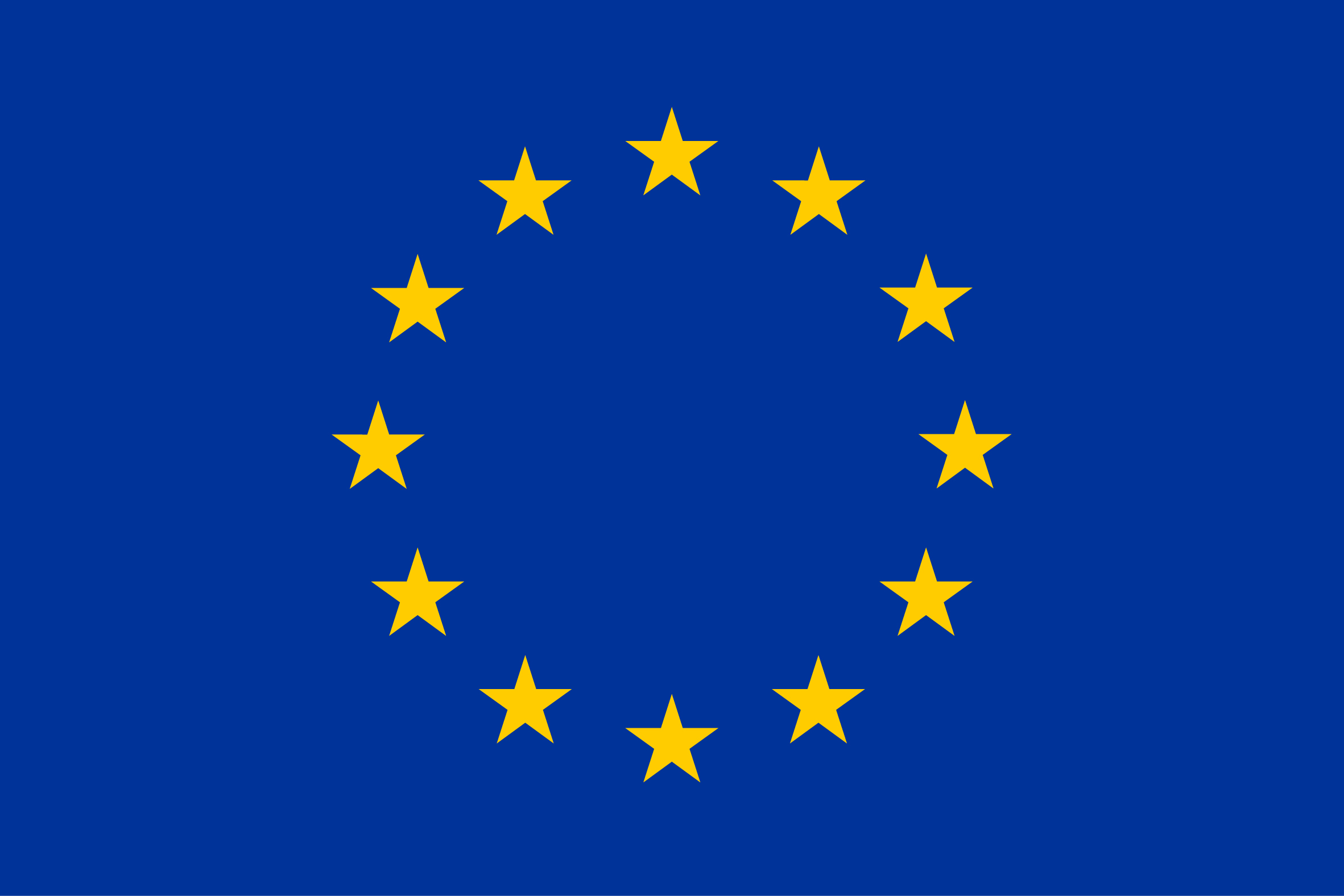}\\
This project has received funding from the European Union’s Horizon 2020 research and innovation programme under the Marie Skłodowska-Curie grant agreement No 860830
\end{center}

\end{document}

%% file: math_commands.tex
\usepackage{amsmath,amsfonts,bm}

\def\eqref#1{equation~\ref{#1}}

\def\1{\bm{1}}

\DeclareMathAlphabet{\mathsfit}{\encodingdefault}{\sfdefault}{m}{sl}
\SetMathAlphabet{\mathsfit}{bold}{\encodingdefault}{\sfdefault}{bx}{n}

